\documentclass[runningheads]{llncs}

\usepackage{float}
\usepackage{graphicx}
\usepackage{tabularx}
\usepackage{eccv}

\usepackage{eccvabbrv}

\usepackage{graphicx}
\usepackage{booktabs}

\usepackage[accsupp]{axessibility}  

\usepackage[pagebackref,breaklinks,colorlinks,citecolor=eccvblue]{hyperref}

\usepackage{orcidlink}

\begin{document}

\title{WorldGPT: A Sora-Inspired Video AI Agent as Rich World Models from Text and Image Inputs} 

\titlerunning{WorldGPT}

\author{%
\begin{tabular}{ccc}
Deshun Yang\thanks{Authors contributed equally.} & Luhui Hu\textsuperscript{$\star$} & Yu Tian \textsuperscript{$\star$} \\
\textit{Seeking AI} & \textit{Seeking AI} & \textit{Harvard University} \\
 & & \email{ytian11@meei.harvard.edu} \\[2ex]
Zihao Li & Chris Kelly & Bang Yang \\
\textit{Seeking AI} & \textit{Stanford University} & \textit{Peking University} \\ \email{li981354@seeking.ai} & \email{ckelly24@stanford.edu} & \email{yangbang@pku.edu.cn} \\[2ex]
Cindy Yang & Yuexian Zou & \\
\textit{University of Washington, Seattle} & \textit{Peking University} & \\
 \email{c1ndyy@uw.edu} & \email{zouyx@pku.edu.cn} & 
\end{tabular}
}

\institute{} 

\authorrunning{D. Yang, L. Hu, Y. Tian et al.}

\maketitle

\begin{abstract}
Several text-to-video diffusion models have demonstrated commendable capabilities in synthesizing high-quality video content. However, it remains a formidable challenge pertaining to maintaining temporal consistency and ensuring action smoothness throughout the generated sequences. In this paper, we present an innovative video generation AI agent that harnesses the power of Sora-inspired multimodal learning to build skilled world models framework based on textual prompts and accompanying images. The framework includes two parts: prompt enhancer and full video translation. The first part employs the capabilities of ChatGPT to meticulously distill and proactively construct precise prompts for each subsequent step, thereby guaranteeing the utmost accuracy in prompt communication and accurate execution in following model operations. The second part employ compatible with existing advanced diffusion techniques to expansively generate and refine the key frame at the conclusion of a video. Then we can expertly harness the power of leading and trailing key frames to craft videos with enhanced temporal consistency and action smoothness. The experimental results confirm that our method has strong effectiveness and novelty in constructing world models from text and image inputs over the other methods.
\end{abstract}

\section{Introduction}
\label{sec:intro}

Text-to-Image Generation Technology such as DALL-E 2 \href{}{\cite{ramesh2021zeroshot}}, CogView 2\cite{hong2022cogvideo}, Imagen\cite{datta2023prompt},  and Stable Diffusion\cite{rombach2022highresolution} have become the leading exemplars in this field. These models harness advanced deep learning techniques and natural language processing to translate text into corresponding high-quality images. For instance, DAll-E 2 employs a transformer-based language model to interpret text prompts, and combines with the StyleGAN 2\cite{karras2020analyzing} model to produce diverse and realistic images.Another path, Stable Diffusion generates images consistent with text prompts by  converting them through CLIP's\cite{radford2021learning,nichol2022glide} text encoder into conditional information,which control latent image representations. 

Text-to-Video Generation Technology like Sora\cite{liu2024sora} and others developed by research teams, which can autonomously create coherent  video content from textual commands. Such systems typically require more intricate model architectures, which not only comprehend textual context but also generate  sequential frames over time to form seamless video streams.

 Our work aims to accelerate the successful application of diffusion models\cite{fakhravar2022international,roos2022layeradapted} to the video field. Due to the fact that many outperforming multimodal processing models are closed-source. We propose the WorldGPT, which can address a variety of challenges such as 1) Temporal-Spatial Consistency: each frame in a video is tightly interconnected with its preceding and following frames. 2) Diversity and Creativity: a requirement for controlled generation of diverse and varied video frames. 3) Video inference Evaluation: assessing the quality of the generated video content and single frame. Our methods incorporate existing complex methods such as dynamic scene modeling, spatiotemporal prediction\cite{yang2020selfsupervised}, and multimodal fusion empowering users to generate a variety of video sequences from text and image inputs. Fig. \ref{fig:results} shows some existing image diffusion methods.

\section{Related Work}

\subsection{Large Language Model}
LLM\cite{kelly2023unifiedvisiongpt,sutskever2011generating,kaplan2020scaling,raffel2020exploring} technology's evolution has been marked by significant advancements since its inception. LLM is a deep learning model with hundreds of millions or even billions of parameters, mainly used to handle natural language tasks, which is defined as:
\[[ \mathcal{L} = -\sum_{i=1}^{n} \log P(\text{token}{i} \mid \text{tokens}{<i}, \theta) ]\]
where \(\text{tokens}_{<i}\) represents the first \(i-1\) tokens (\ie, the context) seen by the model.\(\text{token}_{i}\) is the actual \(i\)th token.\(\theta\) denotes the set of model parameters.

\subsection{Text-to-Image Generation}
Text-to-Image synthesis refers to the process by which artificial intelligence methods, particularly deep learning, which converts natural language description into image. The evolution of the technology promotes the rapid improvement of text-to-image quality. It is successful for conditional GANs\cite{8727938} to transform paired image data accurately. More Importantly, DALL-E \cite{ramesh2021zeroshot}developed by OpenAI could understand and create a wide range of realistic images from natural language prompts.

\subsection{Image-to-Video Generation}
In our multi-step processing pipeline, a pivotal stage prominently employs the cutting-edge DynamiCrafter technology. It is an advanced image-to-video\cite{Vondrick2016GeneratingVI,Chan2019EveryPV} generation technique that harnesses deep learning models to convert static images into continuous, dynamic video sequences. Specifically, given an input image I and the goal to generate a video sequence of length T, the objective of DynamiCrafter can be mathematically expressed as:
\[ \min_{\mathbf{F}, \mathbf{B}} \sum_{t=1}^{T} L(\mathbf{I}_t, \mathbf{G}(I, \mathbf{F}_t, \mathbf{B}_t)) \]
where  \(\mathbf{F}\) denotes the motion-optical flow\cite{Sun2018PWCNetCF,5551149,2021GMFlow} over time. \(\mathbf{B}\) represents the appearance change field.  \(\mathbf{I}_t\) refers to the ground truth frame of the target video at time step t. \(\mathbf{G}\)is a synthesis function that combines the original image I with the corresponding motion information \(\mathbf{B}_t\) to generate a predicted frame. By minimizing the loss function L, it ensures the time and space consistency of generation video frames.

\subsection{Stable Diffusion}
The Stable Diffusion model is a generative model based on diffusion processes, which has gained widespread application in the field of deep learning, particularly for high-quality image generation tasks. Its fundamental principles can be expounded through several core aspects:

The Stable Diffusion model employs a diffusion process,  starting from a completely random noise image and iteratively approaching the probability distribution of real data through a series of denoising steps. This process can be mathematically represented as a Markov chain from the previous state:
\[ x_t = \sqrt{\alpha_t}x_{t-1} + \sqrt{1 - \alpha_t}\epsilon \]
\(x_t\) denote the state at time step t. \(\alpha_t\)is a diffusion coefficient between 0 and 1. \(\epsilon\) is a noise term sampled from a standard normal distribution.

Then, the core component of the model includes a U-Net \cite{8932614}architecture neural network that plays a pivotal role in the denoising process by estimating the adjustments needed to revert the current noisy image back to the original data.

Additionally the Stable Diffusion model integrates cross-modal models such as CLIP, allowing the model to condition its image generation on given textual descriptions.

In summary, the Stable Diffusion can achieve efficient and stable generation of high-quality images within a low dimension latent space\cite{9878449,9008515}.

\section{WorldGPT}
\subsection{Framework Overview}
\begin{figure}[H]
    \centering
    \includegraphics[width=1\linewidth,height=1\textheight,keepaspectratio]{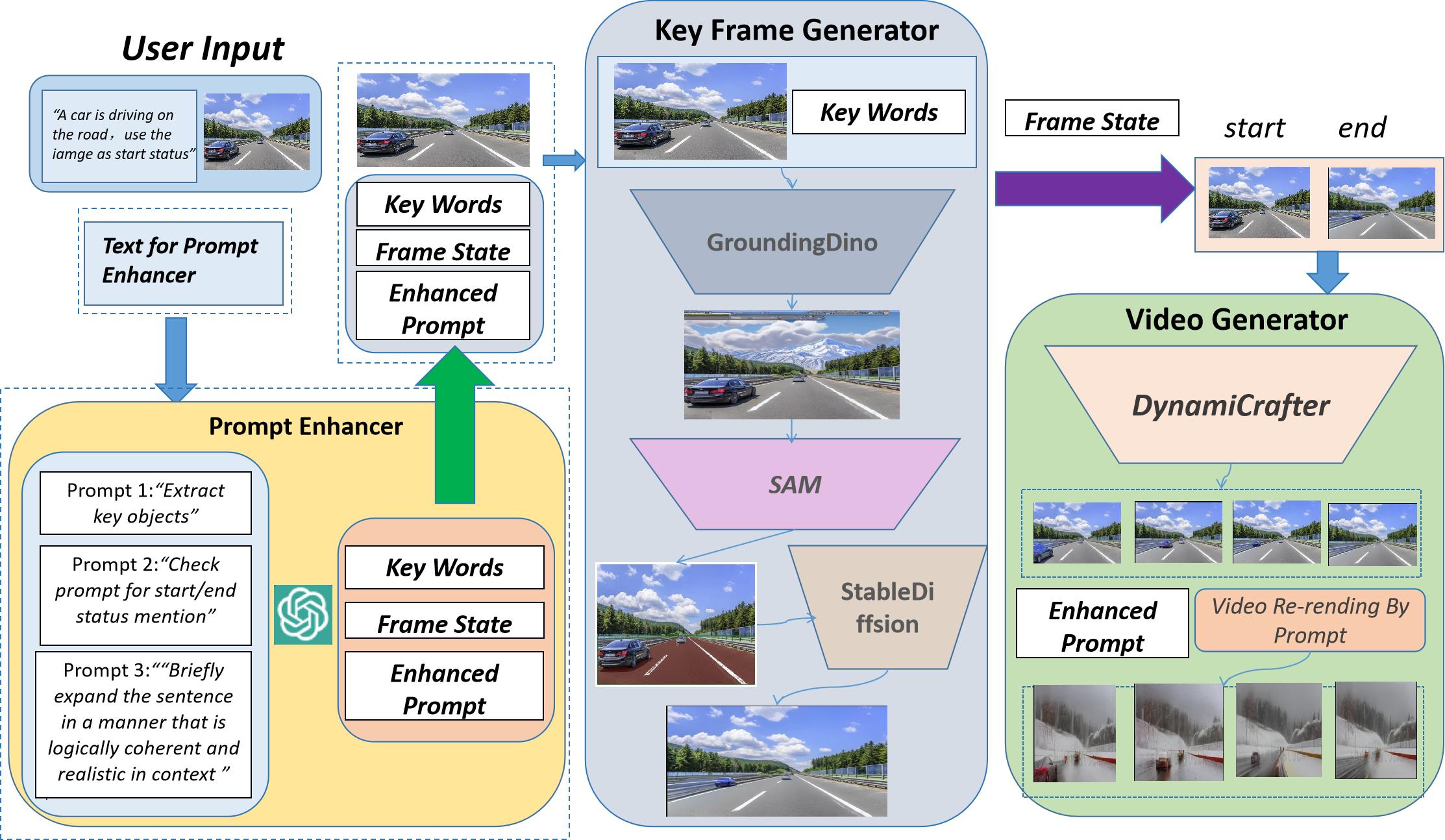}
    \caption{The framework depicted in the figure, follows a concise workflow: Users input Image-Text data, which undergoes transformation by our LLM-based Prompt Enhancer, resulting in three specialized sub-prompts—keywords, input frame state, and optimization prompt. These sub-prompts, alongside the original image, serve as multimodal inputs for the Key Frame Generator. Utilizing a suite of SOTA models, the generator crafts a key frame, which, when paired with the initial image, is then inputted into the Video Generator. With the guidance of the optimization prompt, the Video Generator synthesizes the final video.}
    \label{fig:framework}
\end{figure}

As shown in the Fig.  \ref{fig:framework}, our algorithm is architected with two core components: the Prompt Enhancer and Full Video Translation, which further includes the Key Frame Generator and the Video Generator subsystems. A distinctive feature of our approach is the integration of several leading-edge large models, which enhances our algorithm's ability to generate videos from image-text data without any pre-training, known as zero-shot video generation\cite{9880330}.

This integration not only expands the algorithm's adaptability across various data types but also significantly improves the semantic coherence between the visual content and textual descriptions.

\subsection{Prompt Enhancer}
Using ChatGPT as a prompt enhancer serves the purpose of refining and upgrading user-supplied queries or initial prompts to achieve greater precision, relevance or depth. This capability allows ChatGPT to better grasp user intentions and contextually formulate improved prompts, thereby facilitating more accuracy in prompt communication and accurate execution in following model operations. Given an input sequence(S), the objective is to transform it into a series of task-specific prompt sequence \(P_1,P_2,...,P_n\).
\[\mathcal{L}(\theta;S)=\sum_{i=1}^{n} Loss_i(f(S; \theta), P_i^*) \]
where \(\theta\) represents the set of model parameters.\(f\) is an abstract function representing ChatGPT's process of understanding the input text and generating prompts. \(P_i^*\) is the ideal prompt for the (i)-th subtask.\(Loss_i\) is a loss function measuring the difference between the generated prompt \(f(S; \theta)\) and the ideal prompt \(P_i^*\).

\subsection{Key Frame Generator}
Employing GroundingDino\cite{liu2023grounding} technology to conduct Open-Set object detection on input images with a particular focus on key objectives identified and particular by ChatGPT\cite{10113601}, resulting in meticulous generation of corresponding target masks. Subsequently,  leveraging this critical information,  the Stable Diffusion model is guided to create the final frame for a video, where the task specific \(P_i^*\) prompt driving the generation process is  intelligently distilled by ChatGPT. (I) denote the input image, \(M\) represent the set of masks for detected objects,  \(T\) be the text prompt generated by the ChatGPT, and \(V\) is the synthesized video end frame. Alternatively, we can directly create \(V\) by the following process
\[V = L_{detect}(I, M_{key}) + \lambda \cdot L_{mask}(M_{key}) + L_{video}(T)\]

    Using ChatGPT to generate text prompts of the image content, and applying these prompts to the stable diffusion model to generate video end frames, it achieves an innovative transition from start image of video. This approach not only ensures a high degree of thematic consistency between the video and original image but also provide more artistic creation and flexibility.
    
    In summary, this method harnesses the strengths of cutting-edge visual and language models to deliver accurate object processing while enhancing creative flexibility and adaptability. It is suite for a variety of cases, especially those requiring rapid response to changing contexts,automated content creation, and media storytelling.
    
\subsection{Video Generator}
In the process of building personalized video content using our method, the first step employ leveraging the DynamiCrafter\cite{xing2023dynamicrafter} to generate an seamless sequence of frames based on the provided starting and ending frames. The starting frame signifies the beginning of the video sequence, while the ending frame serves as a definitive blueprint for the desired final state, ensuring that the generated video aligns closely with the user's expectations.
    
Subsequently, to further refine and enhance the video's narrative depth and visual appeal, the initially generated video frames are adjusted by background details and requirements generated by ChatGPT. This deep customization approach allows the final output to maintain the fundamental structure created by the DynamiCrafter, after while incorporating the client's unique vision for background elements, storytelling,and other aspects.

Through this integrated series of steps, the workflow achieves a fluid transformation from concept to reality. The method of create high-quality video content follows both the preset ending and the individual expression. Through this series of exquisite integration and innovative design, we can successfully meet the diversified needs of clients.

\section{Experimental Results}

\subsection{Implementation Details}
In this section, we will conduct evaluation experiments on our agent-based WorldGPT algorithm, using a single NVIDIA 4090 GPU. Additionally, we will compare it with other state-of-the-art image-text video generation algorithms to assess its performance. The evaluation encompassed both quantitative and qualitative assessments. In the quantitative evaluation, we analyzed aspects such as video coherency, alignment with input text, and overall video quality. On the other hand, the qualitative evaluation focused on showcasing the controllability of our algorithm in the image-text-to-video approach.

It is worth noting that all our experiments were conducted under a zero-shot setting, meaning that our algorithm was evaluated without any specific fine-tuning or additional training on the specific tasks at hand. This zero-shot approach demonstrates the generalization ability of our algorithm and highlights its potential for broader application domains. Both our proprietary model and the state-of-the-art (SOTA) models used for comparative experiments were trained using a 256-scale architecture and subsequently deployed for inference. This approach ensured that the models were operating at the same scale during evaluation, allowing for a fair comparison of their respective performance and capabilities.

\subsection{Experimental results and analysis}
\textbf{Metrics.}   In our study, we employ AIGCBench as a benchmark to evaluate the performance of our image-text to video generation models. This approach allows us to leverage our algorithm's capability to process multimodal input from images and text, enabling a comprehensive evaluation of video generation quality from both visual and textual perspectives. AIGCBench stands out for its diverse datasets and advanced evaluation metrics. Our focus is on four key metrics: Control-Video Alignment, Motion Effects, Temporal Consistency, and Video Quality. These metrics play a crucial role in analyzing the capability and performance of our algorithm. They allow us to evaluate aspects such as temporal-spatial consistency (ensuring tight interconnection between each frame in a video and its preceding and following frames), video quality, and semantic coherence, thereby providing us with a comprehensive assessment of our algorithm's performance. 

In our experiments, we compared our algorithm with state-of-the-art Image-Text-to-Video models, such as DynamiCrafter and I2VGen-XL\cite{10113607}, across the four evaluation dimensions mentioned above. The aim was to assess our algorithm's performance in terms of video continuity and alignment with input images. From the results presented in the Table \ref{table-1}, it can be observed that our algorithm outperforms the compared models, DynamiCrafter and I2VGEN-XL, in the aspects of control-video alignment, motion effects, and temporal consistency. However, it is noteworthy that our algorithm performs relatively poorer in frame quality compared to the other two algorithms. This might be attributed to the impact of image quality during the generation of key frames by our algorithm.

\begin{table}
    \centering
    \caption{Quantitative comparisons with SOTA Image-and-Text video generation model on AIGCBench for zero-shot generation}
    \label{table-1}

    \newcolumntype{L}{>{\raggedright\arraybackslash}p{0.16\textwidth}}
    \newcolumntype{R}{>{\raggedright\arraybackslash}X}
    \newcolumntype{M}{>{\raggedright\arraybackslash}p{0.1\textwidth}}

    \begin{tabularx}{\textwidth}{LRRRR}
        \hline 
        \textbf{Dimensions} & \textbf{Metrics} & \textbf{DynamiCrafter} & \textbf{I2VGen-XL} & \textbf{Ours} \\ \hline
        Control-video Alignment & MSE (First) ↓ & 3515.32 & 4178.43 & 1846.52 \\
        & Image-GenVideo Clip ↑ & 0.843 & 0.820 & 0.883 \\
        & GenVideo-Text Clip ↑ & 0.24 & 0.26 & 0.307 \\
        Motion Effects & GenVideo-RefVideo Clip (Corresponding frames) ↑ & 0.780 & 0.750 & 0.810 \\
        Temporal Consistency & GenVideo Clip ↑& 0.980 & 0.960 & 0.992 \\
        & GenVideo-RefVideo Clip ↑& 0.780 & 0.750 & 0.792 \\
        Frame Quality & DOVER ↑ & 0.530 & 0.495 & 0.521 \\
        & GenVideo-RefVideo SSIM ↑ & 0.380 & 0.290 & 0.374 \\ \hline
    \end{tabularx}
\end{table}
Our Qualitative Evaluation section highlights the controllability of our algorithm. We compared our algorithm with DynamiCrafter using customized real-world datasets and images from AIGCBench. By controlling the generation process through prompts, we assessed the controllability and flexibility of both algorithms. Visualized results showed that our algorithm outperforms DynamiCrafter in video generation in terms of controllability and flexibility. 
\begin{figure}[H]
    \centering
    \includegraphics[width=0.8\linewidth,height=0.75\textheight,keepaspectratio]{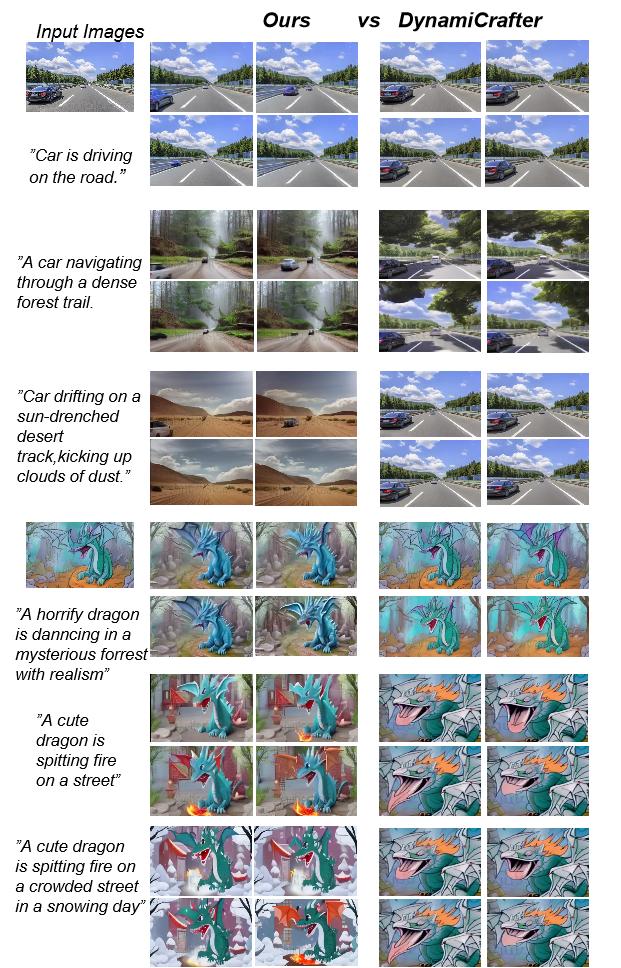}
    \caption{In our qualitative analysis, we input image-text pairs into both our algorithm and the DynamiCrafter algorithm for experiments. The figure depicts this process, showing two sets of image-text pairs on the left. Each set includes an image and three prompts. For every prompt, both algorithms generate a corresponding video. Also, the figure displays four frames located in the top-left, top-right, bottom-left, and bottom-right corners, illustrating the sequence of frames in the generated videos.}
    \label{fig:results}
\end{figure}
According to the Fig. \ref{fig:results} depicted above, our algorithm demonstrates an advantage in terms of alignment with textual input compared to the DynamiCrafter. Furthermore, our algorithm exhibits superior performance in handling complex textual inputs, particularly in generating targets that are not present in the image data. Conversely, when dealing with simple textual inputs, our algorithm may not exhibit perfect alignment subjectively with the original input image. For instance, when generating the third prompt for the first image, DynamiCrafter may exhibit significant differences between the generated image and the textual description due to variations in the content of the prompt compared to the original image. Consequently, the generated video closely resembles the one generated from the first prompt. In contrast, our algorithm demonstrates a comparatively better reflection of the scene described in the textual input. Similarly, when handling prompts for the second image that deviate significantly from the original image, DynamiCrafter fails to generate images that align well with the textual description, resulting in two nearly identical videos. On the other hand, our algorithm's generated videos effectively capture the distinctions and connections between the two prompts.

\textbf{Human Evaluation.}    To further evaluate our algorithm, we conducted human evaluation tests to compare its performance with other state-of-the-art (SOTA) video generation algorithms, focusing on human preference. A total of 50 sets of image-text data were selected from the AIGCBench Dataset, encompassing diverse scenes, styles, and objectives. In pairwise model comparisons, experimenters assessed three perspectives: Text-Video Alignment, Motion Quality, and Visual Quality. The experimental results, presented in Table \ref{table-2}, indicate that our algorithm exhibits comparable visual quality to DynamiCrafter and I2VGen-XL. However, our algorithm demonstrates a significant advantage in terms of motion quality and text-video alignment, with over 60\% of participants choosing our algorithm in the comparisons conducted during the experiments.
\begin{table}
    \centering
\caption{Human evaluation results.The numbers represent the proportion of experiment participants who chose our model.}
\label{table-2}
    \begin{tabular}{cccc} \hline 
         Model Pair&  Visual Quality&  Motion Quality&  Text-Video Alignment\\ \hline 
         Ours vs DynamiCrafterer&  51.3\%&  66.8\%&  62\%\\  
         Ours vs I2VGen-XL&  55.1\%&  72.1\%&  68.7\%\\ \hline
    \end{tabular}

\end{table}
\section{Conclusion}
WorldGPT utilizes a comprehensive pipeline that seamlessly integrates high-level semantic understanding from text with advanced visual interpretation. This innovative approach enables the generation of captivating videos that encapsulate rich and realistic world models. What sets our work apart is the exceptional controllability offered by our image-text algorithm, allowing users to precisely manipulate and shape the video generation process through prompts. This controllability empowers users to customize and tailor the results according to their specific needs and preferences. In summary, WorldGPT pushes the boundaries of AI systems in multimedia generation by combining language and vision, offering both impressive output and user controllability.

\bibliographystyle{splncs04}
\bibliography{main}
\end{document}